# Using Zone Inflation and Volume Transfer to Design a Fabric-based Pneumatic Exosuit with both Efficiency and Wearability

Chendong Liu,[1] Dapeng Yang,[1,*] Jiachen Chen,[1] Yiming Dai,[1] Li Jiang,[1] Shengquan Xie,[2,*] Hong Liu[1]

[1]State Key Laboratory of Robotics and System, Harbin Institute of Technology, Harbin, China

[2]School of Electronic and Electrical Engineering, University of Leeds, Leeds, United Kingdom

E-mail: yangdapeng@hit.edu.cn

E-mail: s.q.xie@leeds.ac.uk

**Keywords:** exosuit, exoskeleton, fabric-based, pneumatic actuator, knee extension assistance

## Abstract

Fabric-based pneumatic exosuits have a broad application prospect due to their good human-machine interaction performance, but their structural design paradigm has not yet been finalized and requires in-depth research. This paper proposes the concepts of zone inflation and volume transfer for the design of a fabric-based pneumatic exosuit with both efficiency and wearability. The meaning of zone inflation is to divide the inflation area of pneumatic exosuit into inflation-deflation zone and inflation-holding zone which can reduce the consumption of compressed air and improve efficiency. Volume transfer, a strategic distribution method of inflatable regions inside the garment, can effectively enhance the wearability of the exosuit. Using inexpensive thermoplastic polyurethane film and clothing fabric, the exosuit is made by heat pressing and sewing. The exosuit has a response time of 0.5s, a stress area of 1500mm$^2$, and a profile of only 32mm, which can be hidden inside common clothing. A mathematical model is developed to predict the output torque of the exosuit with an error of 3.6%. Mechanical experiments show that the exosuit outputs a torque of 9.1Nm at a pressure of 100kPa. Surface electromyography experiments show that the exosuit can provide users with a boost from sitting to standing, with an average reduction in electromyography signals of 14.95%. The exosuit designed using these methods synthesizes efficiency and wearability and is expected to be an ideal paradigm for fabric-based pneumatic exosuits.

## Introduction

Exoskeletons are mechanical devices used for human assistance and are divided into two main types: rigid exoskeletons and flexible exosuits.[1-3] Exosuits are mainly made of soft materials and have better human- machine interaction properties than exoskeletons.[4-6] Currently exosuits can be categorized into contraction or expansion type according to the different driving methods.[7,8] The contraction type is mainly driven by using ropes, pneumatic muscles, shape memory alloys, etc., and has shear force problems.[9] The expansion type mainly uses pneumatic actuators to apply a positive pressure to the human body, which can effectively avoid shear force problems and bring a more comfortable wearing experience.[9]

The fabric-based pneumatic exosuit is of particular interest among the expanding type exosuits. The fabric material used for the exosuit is commonly used in the clothing industry.[10] This makes the exosuit light and soft, with a comfort level comparable to that of clothing, and thus shows great potential in the field of body-assisted devices. A number of representative fabric-based pneumatic exosuits have emerged. Walsh et al. developed the exosuit for shoulder joint assistance, which is equipped with a portable power source in addition to the actuator and connecting parts made of fabrics. It has shown significant effects in medical rehabilitation and industrial assistance.[10-16] The exosuit developed by Zhang et al. can provide assistance for knee extension and flexion.[17-20] Fang et al. proposed a fabric-based pneumatic exosuit with an accordion-like structure that can provide support for the knee joint.[21] Veale et al. proposed a high torque fabric-based pneumatic exosuit.[22] Yeow et al. proposed a fabric-based pneumatic exosuit for the hip joint.[23,24] Many other researchers have designed fabric-based pneumatic exosuits for hand rehabilitation.[25-27]

The design of fabric-based pneumatic exosuits still has some issues that need to be addressed. On the one hand, there is the problem of how to improve the efficiency of the pneumatic system and increase the speed of response so that the exosuit can keep up with faster body movements.[9] On the other hand the problem is how to make the exosuit more clothing-like in order to fully utilize its advantages in wearability.[28] For the first problem there are two solution ideas. One is to improve the performance of pneumatic components, such ac air pumps, but this may bring about an increase in the volume and weight of pneumatic power systems, as well as an increase in the cost of the entire device. The other idea is to do something from the structure of the exosuit. In many cases we don't need to inflate the entire actuator when we need a boost and deflate the entire actuator when we need to unload. We only need to inflate a partial zone of the actuator and deflate a partial zone of it. To the best of our knowledge, there are three works that use this zone inflation method, Park et al. and Sridar et al. for the knee joint and Walsh et al. for the elbow joint.[9, 19, 29] The exosuit developed by Sridar et al is inflated and deflated only in the partial zone of the knee joint, while rigid 3D-printed materials are used in the thigh and calf.[19] This compromises the biggest advantage of the exosuit, softness, but the problem is that the forces generated by the exosuit in the joint zone cannot be transmitted to the limb without the use of high stiffness connectors. In fact high stiffness can be achieved just as well with inflatable soft materials. This is the case with the work of Park et al.[29] and Walsh et al.[9] The method of zone inflation used in this paper is inspired by these two works, and we combined it with our previously proposed method of volume transfer [28] that can effectively enhance the wearability of the exosuit, resulting in a fabric-based pneumatic exosuit with both efficiency and wearability. This structural form of exosuit performs satisfactorily in various indicators such as the output torque, response time, profile size, and stress area, and we believe that it is expected to be an ideal paradigm for fabric-based pneumatic exosuits.

Specifically, this fabric-based pneumatic exosuit is fabricated using common apparel fabrics, which are light, soft, and inexpensive, and provide assistance for knee extension. The entire exosuit is divided into the inflation-deflation zone and the inflation-holding zone. The inflation-deflation zone is located in the joint area. This zone is inflated to provide a boost to the joint and deflated to unload. The inflation -holding zone is located in the limb area and is inflated as soon as the device is activated and maintains a high degree of stiffness throughout use, thus transmitting the forces generated by the inflation-deflation zone to the limb. The use of this zone inflation method effectively reduces the amount of gas change required for joint movement and improves the pneumatic efficiency, resulting in a fast response with a response time of only 0.5s. Using our previously proposed method of volume transfer, arranging the inflatable area inside the garment effectively reduces the profile of the exosuit (only 32 mm) so that it can be hidden inside common garments, and increases its stress area (1500 mm$^2$) to improve comfort. The result of the combined use of the two design methods is the birth of an exosuit with both pneumatic efficiency and wearability, which promises to be an ideal paradigm for fabric-based pneumatic exosuits. This paper describes in detail the design concept and manufacturing method of this exosuit. A mathematical model was developed to describe it, and its output torque and response time were obtained through static and dynamic experiments. Its assistance effect was tested by experiments with electromyographic signals of knee movements from sitting to standing.

## Exosuit Designed with Zone Inflation and Volume Transfer

The design concept of zone inflation and volume transfer is shown in Figure 1. The exosuit designed is shown in Figure 1a. The exosuit is sewn from fabrics, light and soft. The entire exosuit is divided into the inflation-deflation zone and the inflation-holding zone. The inflation-deflation zone is represented by red dashed lines, and the inflation-holding zone is represented by white dashed lines. The inflation- deflation zone is arranged near the joints, and there are changes in inflation and deflation during movements. When inflating, it corresponds to high stiffness and outputs torques, while when deflating, it corresponds to low stiffness and achieves unloading. The inflation-holding zone is arranged on the limbs, inflated before the start of movements, and holds a rigid state throughout the entire movement process, so as to effectively transmit the assist torque generated by the inflation- deflation zone to the limbs. After the movement stops completely, this zone deflates and returns to a soft state. The use of this zone inflation design concept can effectively reduce the volume of compressed air required to be supplied during exosuit movements, thereby reducing the burden on the pneumatic system and achieving higher response speed, while avoiding the use of rigid components and maintaining the softness and comfort of exosuits.

Figure 1b shows the effect of wearing the exosuit. Figure 1c is a cross-sectional view of the leg, demonstrating the design concept of volume transfer. Volume transfer is a design concept proposed in our previous work to improve the wearability of exosuits.[28] The traditional exosuit design uses two cylindrical actuators (black dashed line in Figure 1c). In this work, we used the method of volume transfer to adjust the arrangement of the actuators to four cylindrical actuators (brown lines in Figure 1c). This can effectively reduce the profile size of the exosuit, increase the stress area of the exosuit, while maintaining sufficient output torques. The detailed explanation of volume transfer can refer to our previous work.[28] In this work, we did not use 7 actuators as before,[28] but reduced them to 4 because after using zone inflation, a certain amount of space needs to be left for arranging pneumatic pipelines. Figure 1d shows the excellent wearability of this exosuit, which can be hidden inside common clothing.

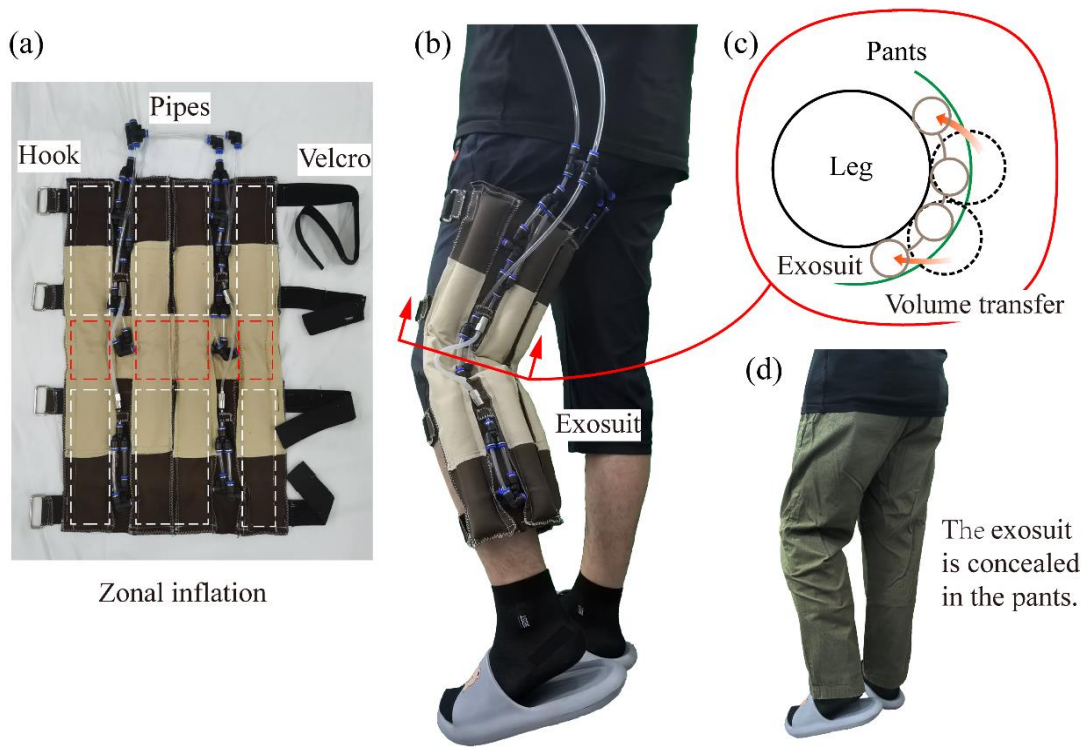

Figure 1. The exosuit designed with zone inflation and volume transfer. (a) The fabric-based pneumatic exosuit designed by this work. Made from commonly used clothing fabrics in the textile industry, it is soft and comfortable to wear. Adjustable binding is achieved through hooks and velcro. The exosuit is divided into one inflation-deflation zone (red dashed area) and two inflation-holding zones (white dashed area). (b) A user wearing the exosuit. (c) A cross-sectional view of the leg, used to represent the concept of volume transfer. Brown represents the exosuit. Green represents pants. The arrow indicates the direction of volume transfer. The black dashed line represents the traditional design that does not use volume transfer. (d) The exosuit's profile size is small enough so that it can be hidden inside common clothing.

## Fabrication

Figure 2 shows the fabrication process of the exosuit. We use common pure cotton fabrics for fabrication. This kind of fabric is widely used in the fabrication of various coats and trousers in the clothing industry. It is light and soft and very comfortable to wear. Step 1 is to cut a piece of yellow fabric (No. 1) as the base fabric. Step 2 is to cut two pieces of brown fabrics (No. 2 and 3) and put them on the No. 1 fabric. Step 3 is to cut 8 long brown fabrics (No. 4 to 11) and put them on the No. 2 and No. 3 fabrics. Step 4 is to cut 4 long yellow fabrics (No. 12 to 15) and put them on the No. 4 to 11 fabrics. All the fabrics required until this step have been cut and arranged.

Because pure cotton fabrics used in the clothing industry are breathable, additional measures need to be taken to achieve sealing. The materials used for sealing should be as light and soft as cotton fabrics so that the wearability of the exosuit is not compromised. The vacuum bag film made of thermoplastic polyurethane (TPU) is an ideal material. This material is very thin and has good toughness. The material type used in this work is Stretchlon 200 (Fiberglass, USA). Using this material to fabricate sealed airbags and stuffing them into the fabrics, the deformation of the airbags can be limited by the high strength of the fabrics, so that the exosuit can withstand higher air pressures. Step 5 is to fabricate two different types of sealed airbags. Type a is fabricated by hot pressing two films. Type b is fabricated by hot pressing 4 layers of films. The hot pressing tool used in this work is an iron with a temperature of 160 °C. A total of 8 type a airbags and 4 type b airbags were fabricated. Step 6 is to stuff the hot pressed sealed airbags into the fabrics. The stacking sequence of the sealed airbags and fabrics can be shown in the cross-sectional view A-A. The direction of the cross-sectional view is marked in step 4. The numerical numbers on the cross-sectional view represent the fabrics corresponding to the numbers in step 1-4. The black arrows correspond to the two types of sealed airbags in step 5. The black dashed line represents the position where the fabrics are sewn. Step 7 is to sew the fabrics to wrap the sealed airbags inside the fabrics. Sewing according to the path (black dashed line) can divide the inflation-deflation zone and the inflation-holding zone. Step 8 is to sew the hooks, elastic bands, and velcro for binding, connect the corresponding pneumatic pipes, and the exosuit is finished.

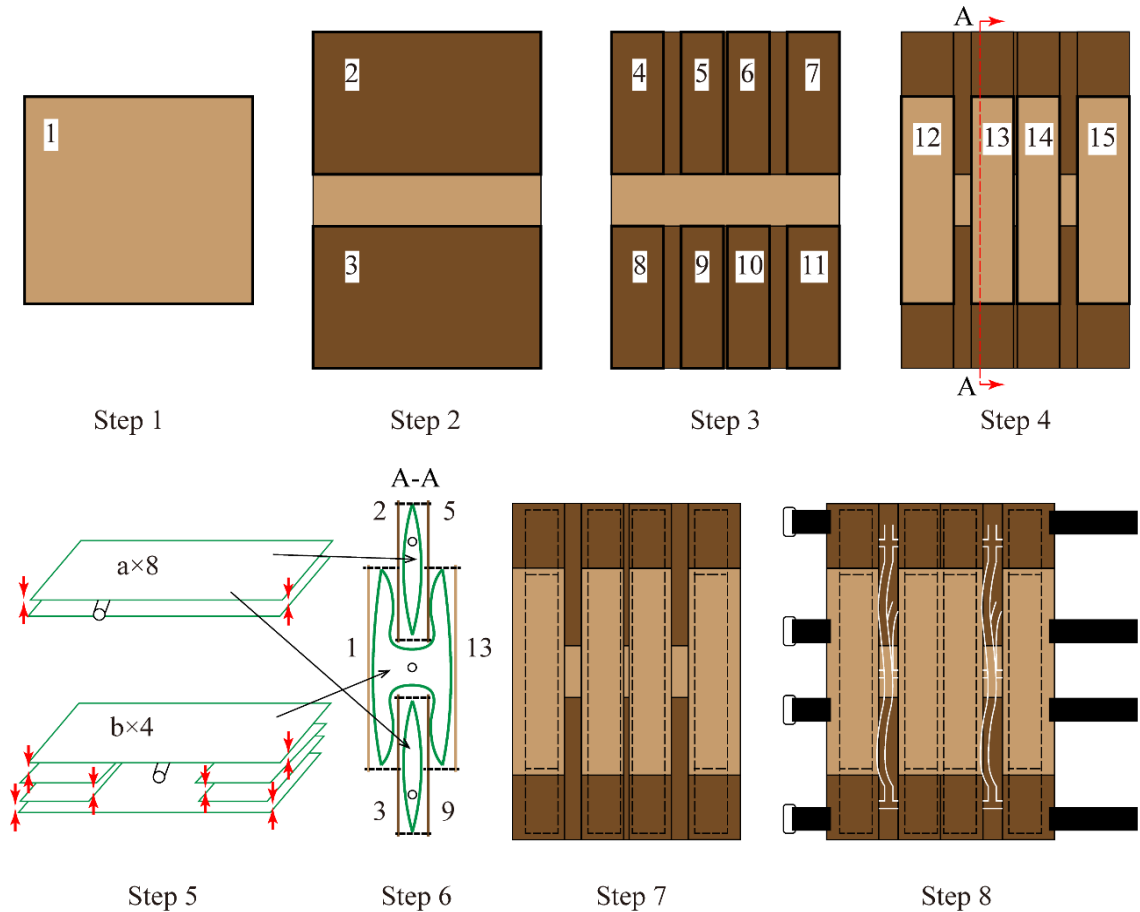

Figure 2. The fabrication process of the exosuit is divided into 8 steps. Steps 1 to 4 are the cutting and sequential arrangement of fabrics. A total of 15 fabrics of different sizes were used, and for ease of explanation, each fabric was labeled with a numerical number. Step 5 is the hot pressing fabrication process of sealed airbags. Two types of sealed airbags were fabricated, 8 in type a and 4 in type b. The red arrow indicates the hot pressing direction. Step 6 inserts the sealed airbag between the fabrics. For ease of explanation, a cross-sectional view is provided. The direction of the cross-sectional view is marked in step 4. The numerical numbers on the cross-sectional view represent the fabrics corresponding to the numbers in step 1-4. The black arrows correspond to the two types of sealed airbags in step 5. The black dashed line represents the position where the fabrics are sewn. Step 7 represents the path of the sewing thread. Step 8 adds pneumatic tubing (white line), hooks and velcro for binding.

## Mathematical Model

We have established a mathematical model based on the principle of virtual work in our previous work,[28] which can be used to guide the design of exosuit with volume transfer design method and predict its output torques. Due to the addition of zone inflation in the exosuit designed in this work, certain adjustments need to be made in the mathematical model.

The main changes are shown in Figure 3a. Due to the division of the previous complete cylindrical actuator into three zones, namely the inflation-deflation zone in the middle and the inflation-holding zones at both ends, certain size limitations need to be imposed on each region. The general principle is to ensure that only the volume of the inflation-deflation zone changes when the actuator is bent, while the volume of the inflation-holding zone remains unchanged. The purpose of this is to make the assist torque completely generated by the inflation-deflation zone, so that we don't need to control the air pressure in the inflation-holding zone (just give a fixed value). In order to achieve the above purpose, Formula (1) needs to be satisfied:

$$l > 2d \tan(\frac{\theta}{2}) \tag{1}$$

The meanings of the variables in the formula are shown in Figure 3a, and $l$ represents the length of the inflation-deflation zone, $d$ represents the diameter of the cylindrical actuator, $\theta$ represents the bending angle of the cylindrical actuator. By visualizing Formula (1) as Figure 3b, the relationship between the three variables can be intuitively observed. The value of $l$ selected during design process should be above the surface in Figure 3b.

To know the output torque of the exosuit, Formula (2) can be used:

$$T = \frac{\pi n p d^3}{8\cos^2(\frac{\theta}{2})} \tag{2}$$

In the formula, $T$ represents the output torque of the exosuit, $n$ represents the number of cylindrical actuators, and $p$ represents the internal air pressure of the actuators. The detailed derivation of this formula can refer to our previous work.[28] Figure 3c is the parameter design diagram of the exosuit. This diagram is the visualization result of Formula (2). From the diagram, it can be visually observed how the use of different numbers and diameters of cylindrical actuators will affect the final output torque. The corresponding air pressure $p$ in this graph is 100kPa, and the bending angle $\theta$ of the exosuit is 80°. If only wearability is considered, smaller $d$ and larger $n$ can bring lower profile size and larger stress area. But we also need to consider the possibility of fabrication process and the layout of pneumatic pipelines. Considering the mechanical properties, wearability and fabrication process, we chose the star point in the diagram as the final design. The diameter of the cylindrical actuator is 32mm, the number of actuators is 4, and the expected maximum output torque is 8.77Nm. In our previous work, we also used a similar parameter design diagram. At that time, we decided that the number of actuators $n$ was 7.[28] The decrease of $n$ from 7 to 4 this time is due to the addition of zone inflation, which caused the pneumatic pipeline to occupy a certain amount of space. After determining the design parameters of the exosuit, the mathematical model can be used to predict the output torque of the exosuit under different air pressures $p$ and bending angles $\theta$, as shown in Figure 3d.

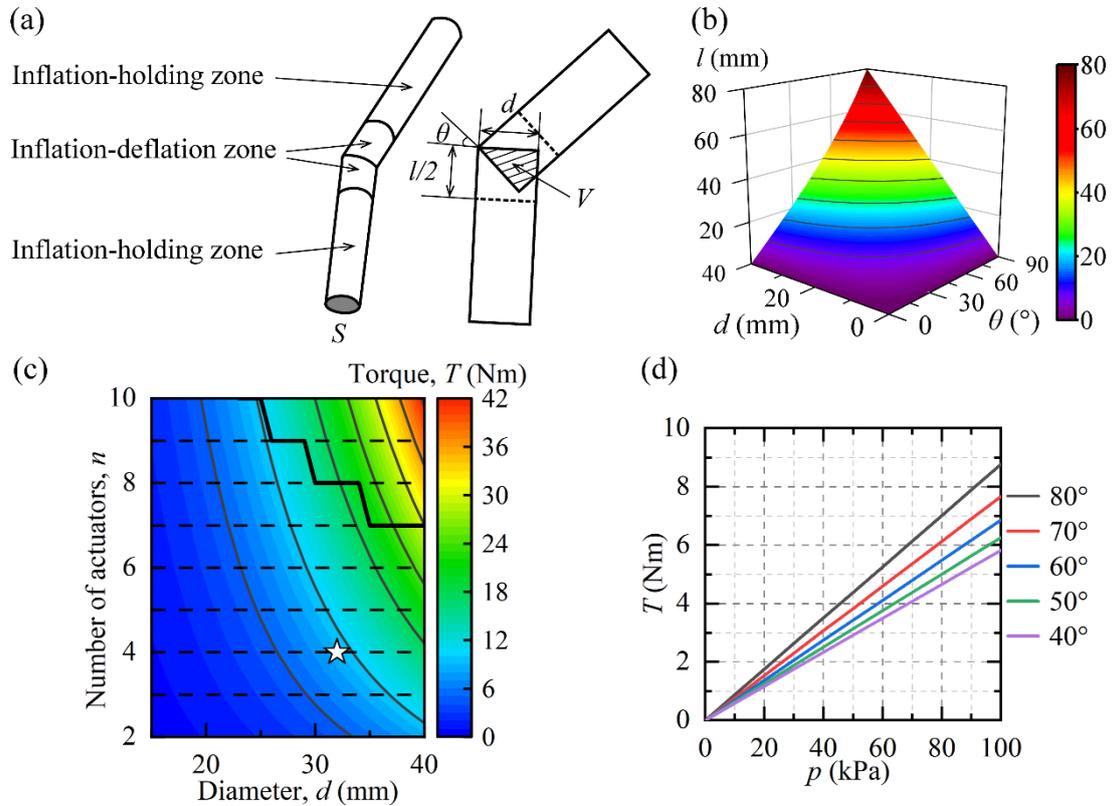

Figure 3. Mathematical model used to guide design and predict output torque. (a) Schematic diagram of main parameters involved in the mathematical model. (b) Relationship between the diameter $d$ of the cylindrical actuator, the bending angle $\theta$ of the cylindrical actuator, and the length of the inflation-deflation zone $l$. (c) Parameter design diagram of the exosuit. (d) Theoretical prediction results of exosuit output torques.

## Static Mechanical Experiment

This section tests the output torque of the exosuit through static mechanical experiments. The experimental equipment used are shown in Figure 4a. We made two 3D printed cylindrical molds to simulate human legs. The diameter of the cylindrical mold is

-9-

approximately equal to the diameter of the human leg. The two molds were installed on the aluminum profile beams. Aluminum profile beams were connected by hinges and can rotate, simulating the movement of knee joints. The exosuit was tied to the 3D printing molds. A protractor was used to measure the rotation angle of the aluminum beam. This angle is approximately equal to the exosuit bending angle $\theta$. Two pressure reducing valves (IR2010-02-A, SMC, Japan) were provided for the inflation-deflation zone and the inflation-holding zone respectively to control the pressure accurately. Two switch valves were used to control the inflation and deflation in the inflation-deflation zone. A force sensor (DYLF-102, DAYSENSOR, China) was used to measure the output force at the end of the aluminum beam, and the output torque of the exosuit can be obtained by multiplying it by the arm of the aluminum beam. During the experiment, the pressure in the inflation-holding zone was maintained at 120kPa. The bending angles $\theta$ were 80°, 70°, 60°, 50°, and 40°.

Figure 4b-f show the relationship between the output torque $T$ and air pressure $p$ at different bending angles $\theta$. Scatters represent experimental measurements, while lines represent mathematical model predictions. From the figure, it can be seen that the mathematical model used in this article can predict the output torque of exosuit very well. The relationship between the output torque and air pressure is approximately linear at different bending angles. The maximum output torque of the exosuit was experimentally measured to be 9.1 Nm. The predicted value of the mathematical model is 8.77Nm. The error between them is 3.6%.

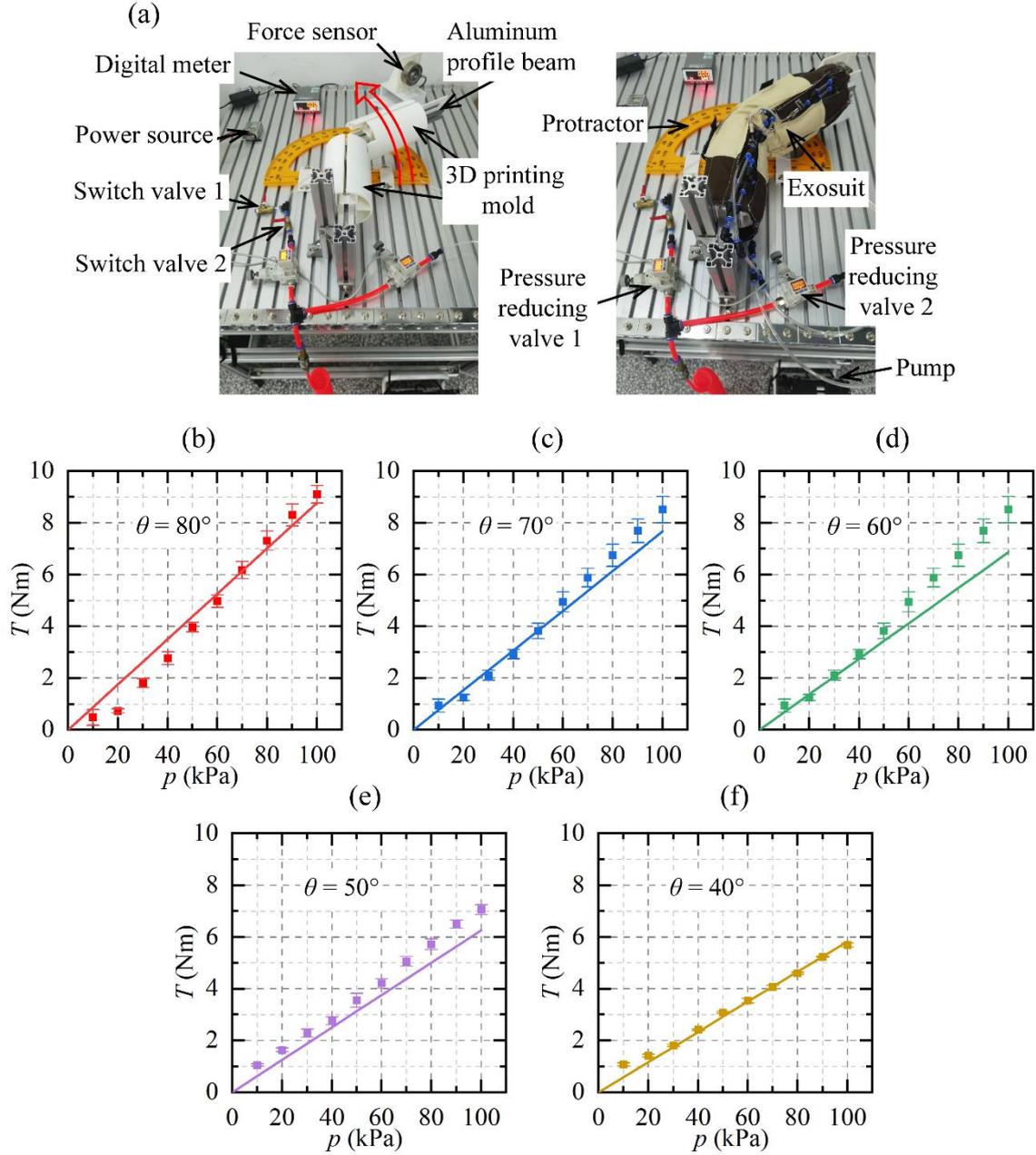

Figure 4. Static mechanical experiment. (a) Experimental equipments. (b)-(f) The relationship between the output torque $T$ and the air pressure $p$ corresponding to the bending angles $\theta$ of 80 °, 70 °, 60 °, 50 °, and 40 °, with scatter points representing measured values and lines representing predicted values from mathematical models.

## Dynamic Response Experiment

This section tests the dynamic response time of the exosuit at target pressures. The target pressures are step signals. The experimental process and equipment used are shown in Figure 5a. An air pump (ED-0204, Eidolon, China) was used to provide compressed air for the entire system. Pressure reducing valve 1 was used to provide the target pressure.

Pressure reducing valve 2 provides a fixed pressure of 120kPa for the inflation-holding zone. Switch valves 1 and 2 were used to control the inflation and deflation of the inflation-deflation zone. The pressure in the inflation-deflation zone was measured in real-time by a pressure sensor (MPX5700GP, NPX, UGA) and input into an Arduino control board, which was ultimately fed back to the computer for processing.

The dynamic response time of four different target pressures (100kPa, 80kPa, 60kPa, 40kPa) was tested in the experiment, as shown in Figure 5b-e. The response times at different target pressures are 0.5s (100kPa), 0.34s (80kPa), 0.32s (60kPa) and 0.3s (40kPa), respectively. The experimental results indicate the exosuit has good dynamic performance.

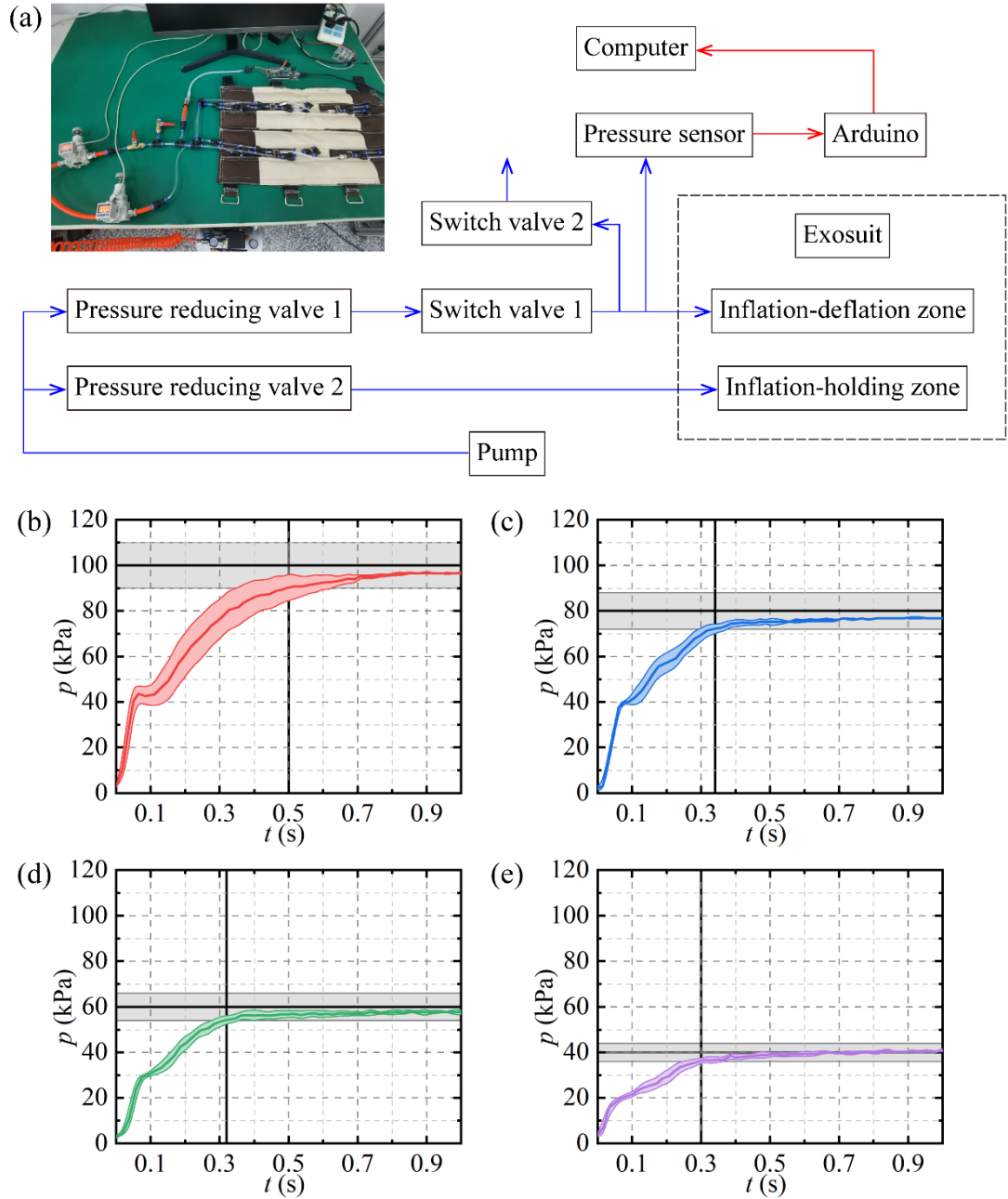

Figure 5. Dynamic response experiment. (a) Experimental processes and equipments. The blue arrow represents the pneumatic circuit. The red arrow represents the electrical signal circuit. (b)-(e) Dynamic response of the exosuit under different target pressures. The shaded part indicates the error range of the target pressure ($\pm 10\%$).

## Surface Electromyography Experiment

This section evaluates the assistive effect of the exosuit on the knee joint by collecting the intensity of the rectus femoris muscle electromyography signal during the sit to stand motion. The experimental process and equipment are shown in Figure 6a. The subjects changed from sitting posture to standing posture. The exosuit can provide assistance for this action, thereby reducing the activity of related muscles (mainly the rectus femoris muscle). The degree of reduction in electromyographic signals can quantitatively reflect the assistive effect of the exosuit.

The wireless surface electromyographys (sEMG) signal sensor (Delsys, USA) was attached to the rectus femoris muscle of the subject's thigh. The inertial measurement unit (IMU) sensor was tied above the knee joint to detect the subject's body posture. An air pump was used to provide compressed air for the whole system. A vacuum pump (V550-30, Tengyuan, China) was used to exhaust air faster. The pressure reducing valve 1 was used to provide the target pressure (100kPa) for the inflation-deflation zone. When solenoid valve 1 was open and solenoid valve 2 was closed, the inflation-deflation zone was inflated. When solenoid valve 1 was closed and solenoid valve 2 was open, the inflation-deflation zone was deflated. The pressure reducing valve 2 was used to provide a fixed pressure of 120kPa for the inflation-holding zone. An Arduino control board was used as the main controller. The information collected by the bluetooth module from the IMU sensor was returned to Arduino for processing. Two relays were used to control the two solenoid valves. The physical display of the equipments is shown in Figure S1 in supplementary materials. A total of 4 subjects were tested, and their relevant information was shown in Table S1 in supplementary materials. The experimental process is shown in Video S1 in supplementary materials.

The test results of sEMG are shown in Figure 6b-e. The electromyographic signals collected by the sensor were first processed by a 4th order Butterworth filter (10-400Hz), and then rectified. By calculating the mean value of sEMG signals within a cycle, we can compare the intensity of sEMG signals before and after using the exosuit. After using the exosuit, the sEMG of the four subjects decreased by 7.7%, 23.2%, 12.9%, and 16% respectively, with an average decrease of 14.95%. The Quebec User Evaluation of Satisfaction with Assistive Technology 2.0 (QUEST) questionnaire[30] was used to measure the 4 participants' satisfaction with the wearable technology (Table S2). The questionnaire is scored from 8 dimensions: dimensions, weight, adjustments, safety, durability, simplicity of use, comfort and effectiveness. Scores from 1 to 5 indicate not satisfied at all, not very satisfied, more or less satisfied, quite satisfied and very satisfied. The results show that participants are quite satisfied with the exosuit. The QUEST total score is 4.28.

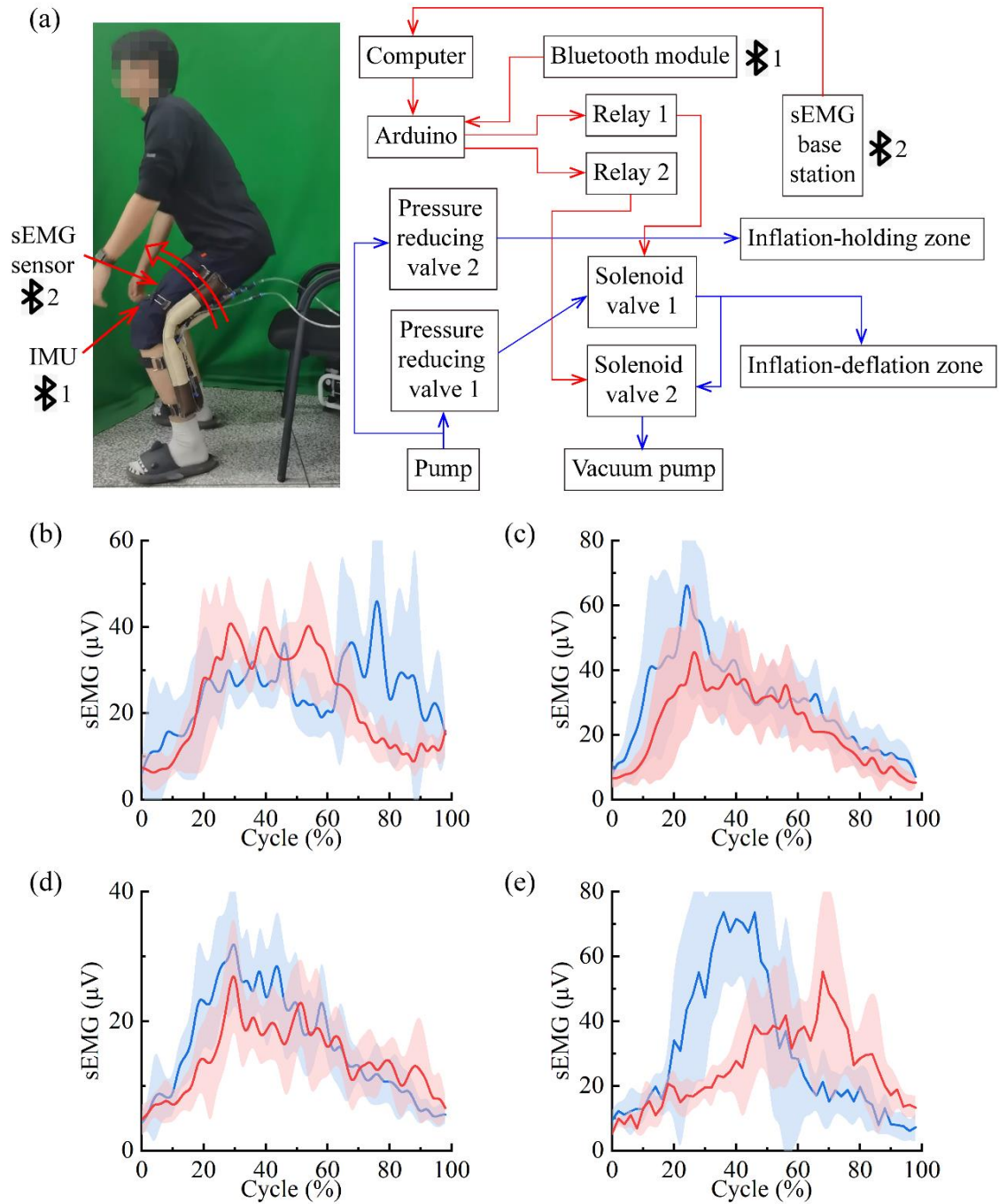

Figure 6. Surface electromyography experiment. (a) Experimental processes and equipments. The blue arrow represents the pneumatic circuit. The red arrow represents the electrical signal circuit. (b)-(e) Surface electromyographic signals of the rectus femoris muscle were collected from four subjects during the sit to stand motion. The blue curve represents the surface electromyographic signal without the exosuit. The red curve represents the surface electromyographic signal with the exosuit. The error band represents the standard deviation.

## Discussion

Table 1 provides a comparison between this work and four other representative works. Comparison indicators include the torque, response time, profile size and stress area. These four indicators can comprehensively reflect the mechanical performance, dynamic performance, comfort, and wearability of the exosuit. It should be declared that it is of little significance to compare a certain indicator separately, because the ideal exosuit paradigm should be the comprehensive optimization of all indicators. The torque indicator is easy to meet. Generally, the torque of more than 6Nm can produce obvious assistance effects [29], so we focus on the three indicators of the response time, profile size and stress area. Ref [19], Ref [29], Ref [9] and this paper used the zone inflation. The response speed reported in these four papers is satisfactory, and the response time is about 1s. Ref [29] does not use zone inflation, so its dynamic performance is poor. Ref [29] and this paper used the volume transfer. Their profile size can be controlled at around 30mm, so they can be hidden inside common clothing. Their stress area can reach about 1/2 of the limb surface. Ref [19], Ref [29], Ref [9] did not use volume transfer, and their profile size is over 60mm, making it difficult to achieve concealment. Moreover, their stress area is small, which can only reach 1/4 to 1/3 of the limb surface.

This work is the first fully soft exosuit designed using a combination of zone inflation and volume transfer, resulting in excellent efficiency and wearability. The ideal exosuit should be lightweight and comfortable like clothing while providing quick assistance, and this work is very close to this ultimate goal. The exosuit presented in this article is intended for the knee joint, and researchers could certainly extend it to other joints as it is not just a device, but a paradigm.

Table 1. Comparison between this work and other representative work.

|  | Torque | Response time | Profile size | Stress area |
| --- | --- | --- | --- | --- |
| Ref [19] | None | 3s | None | ≈1/4 of the leg surface |
| Ref [29] | 12.3 Nm | 0.4s | 60mm | ≈1/4 of the leg surface |
| Ref [9] | 13.5 Nm | 1s | 77mm | 1/4-1/3 of the arm surface |
| Ref [28] | 7.6 Nm | None | 26mm | ≈1/2 of the leg surface |
| This work | 9.1 Nm | 0.5s | 32mm | ≈1/2 of the leg surface (1500mm$^2$) |

# Conclusion

This article uses the design concept of zone inflation and volume transfer to develop a fabric-based pneumatic exosuit that combines efficiency and wearability. By dividing the inflation zone of the exosuit into inflation-deflation zone and inflation-holding zone, the efficiency of pneumatic system is improved, and the response time is reduced (only 0.5s), thus obtaining good dynamic performance. The successful application of the volume transfer method makes this exosuit have good wearability (stress area 1500mm$^2$) and can be easily hidden in common clothing (profile size 32mm). This article provides a detailed introduction to the fabrication method of the exosuit, which can be completed using common low-cost materials and tools. The mathematical model provided can effectively guide the design of the exosuit and predict its output torque, with an error of only 3.6%. The experimental test shows that the maximum output torque of the exosuit is 9.1Nm. The assistive effect of the exosuit on the knee joint during sit to stand motion was evaluated through sEMG experiments, with an average reduction of 14.95% in related muscle activity. The limitation of this article is that the exosuit has a low level of integration and is difficult to apply outside the laboratory. In the future, we will develop a supporting portable system, so that the exosuit can be applied in more scenarios.

We believe that the exosuit proposed in this article is an ideal paradigm, as it comprehensively considers the assistive effect and wearability of wearable devices. This article not only introduces a specific exosuit, but also presents a design method for exosuits. Future researchers can apply this design method to other specific wearable devices, thereby advancing the progress of the entire field.

## Supplementary Material

Supplementary Data
Supplementary Video S1